\documentclass[conference]{IEEEtran}
\IEEEoverridecommandlockouts
\usepackage{cite}
\usepackage{amsmath,amssymb,amsfonts}
\usepackage{algorithmic}
\usepackage{graphicx}
\usepackage{textcomp}
\usepackage{xcolor}
\usepackage{float}
\usepackage[hidelinks]{hyperref}
\def\BibTeX{{\rm B\kern-.05em{\sc i\kern-.025em b}\kern-.08em
    T\kern-.1667em\lower.7ex\hbox{E}\kern-.125emX}}
\begin{document}

\title{JEPTERA: A Platform Agnostic JEPA-Assisted Model Predictive Planner for Urban UAV Navigation}

\author{
\IEEEauthorblockN{
Gibran Fathoni Belva\textsuperscript{1},
Mansur Arief\textsuperscript{2},
Ahmad Alfan Alfian Irfan\textsuperscript{1},
}

\IEEEauthorblockA{
\textsuperscript{1}Department of Information Technology,
Universitas Muhammadiyah Yogyakarta, Indonesia\\
\textsuperscript{2}Department of Industrial and Systems Engineering,
King Fahd University of Petroleum and Mineral, Saudi Arabia\\
\{gibran.fathoni.ft23, ahmad.alfan.ft23\}@mail.umy.ac.id;
mansur.arief@kfupm.edu.sa
}
}

\maketitle

\begin{abstract}
Urban navigation for Unmanned Aerial Vehicles (UAVs) requires planning that accounts for environmental geometry, vehicle dynamics, and moving obstacles. This paper presents JEPTERA, a navigation architecture that integrates a Joint-Embedding Predictive Architecture (JEPA), Model Predictive Control (MPC), PID control, and a safety layer. JEPA predicts the UAV’s motion response to candidate commands for MPC action selection, while the safety layer checks trajectories against occupied space and predicted obstacle motion. Simulator adapters and platform-specific YAML configurations separate the planning core from simulator interfaces. An ablation study evaluated six configurations across three gym scenarios, totaling 180 runs. The complete architecture achieved the highest aggregate success rate, completing 28 of 30 runs. Additional tests completed one route across three city maps without collisions. Separate integration tests demonstrated operation in Gazebo/ROS 2, MuJoCo, and CARLA without modifying the core planning logic. These results support the feasibility of the integrated architecture for simulated urban UAV navigation and its portability across simulator interfaces.
\end{abstract}


\section{Introduction}
 Autonomous UAVs require reliable navigation in dynamic urban environments. Buildings that constrain the airspace, along with moving obstacles, necessitate online planning based on continuously updated observations \cite{zhou2021egoplanner,tordesillas2022faster}. Such conditions demand a planning system capable of accounting for environmental geometry, the UAV's motion constraints, and the predicted movement of obstacles, in order to determine safe maneuvers while still achieving the desired objectives \cite{hewing2020learningmpc}.

\begin{figure*}
    \centering
    \includegraphics[
        width=0.935\textwidth,
        trim=0 0.7cm 0 0.5cm,
        clip
    ]{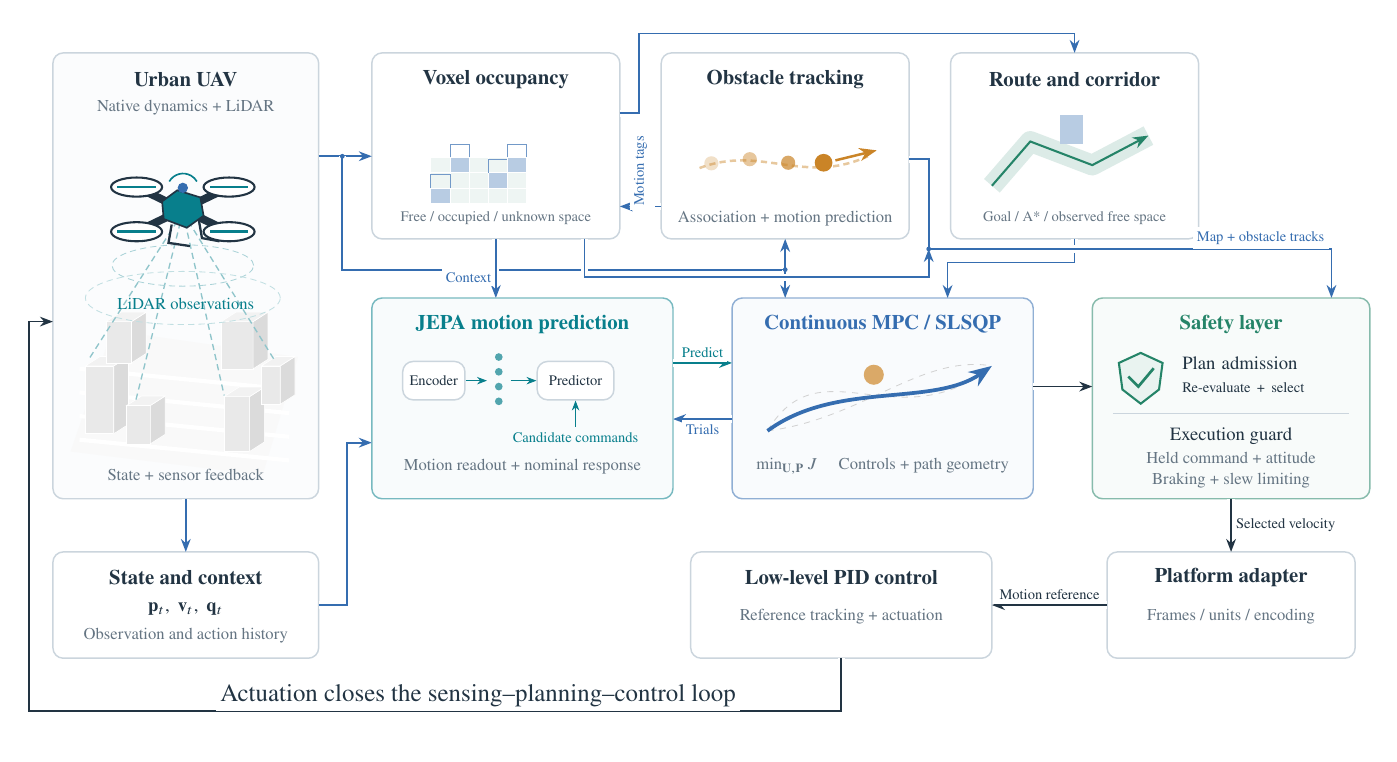}
    \caption{Overview of JEPTERA Architecture}
    \label{fig:jeptera_architecture}
\end{figure*}

Model predictive control (MPC) selects actions by optimizing predicted motion under system objectives and constraints \cite{falanga2018pampc,hewing2020learningmpc}. The Joint-Embedding Predictive Architecture (JEPA) provides a framework for prediction in representation space \cite{lecun2022path}, subsequently explored through I-JEPA for images and V-JEPA 2 for action-conditioned robot planning \cite{assran2023ijepa,assran2025vjepa2}. These developments highlight the potential of representation space prediction for UAV navigation among static and dynamic obstacles.

Although MPC–PID with a safety layer provides reliable navigation, its performance may degrade in complex dynamic environments where anticipating obstacle motion is essential \cite{hewing2020learningmpc}. This limitation motivates integrating JEPA-based motion prediction with MPC to improve navigation reliability under challenging conditions.

This paper proposes \textbf{JEPTERA}, a platform-agnostic UAV navigation
architecture integrating JEPA-based motion prediction, MPC, PID control, and safety layer. Simulator adapters and platform-specific YAML configurations decouple the planning core from platform interfaces. Evaluation includes component ablations, navigation across three urban maps, and cross-simulator integration tests.

 To summarize, our main contributions are threefold.
\begin{enumerate}
    \item a planning architecture combining JEPA-based motion prediction with MPC, PID control, and a safety layer;
    \item a shared planning core integrated across simulators through adapters and platform-specific YAML configurations; and
    \item an evaluation comprising 180 ablation runs, navigation across three urban maps, and cross-simulator integration tests.
\end{enumerate}

\section{Related Work}
\subsection{UAV Trajectory Planning and Predictive Control}
UAV trajectory planning must balance environmental geometry, motion feasibility, and computational efficiency. The Dynamic Window Approach selects local velocities for collision avoidance, while minimum-snap optimization generates smooth quadrotor trajectories \cite{fox1997dwa,mellinger2011minsnap}. Fast-Planner combines kinodynamic search with trajectory optimization for high-speed flight \cite{zhou2019fastplanner}. EGO-Planner reduces planning overhead through ESDF-free gradient-based optimization \cite{zhou2021egoplanner}, while FASTER navigates known and unknown environments with a backup trajectory in verified free space \cite{tordesillas2022faster}.

MPC selects actions using predicted dynamics over a planning horizon \cite{liu2017mpcplanning,jalalmaab2017feasibility}. PAMPC incorporates perception requirements into
quadrotor control \cite{falanga2018pampc}, while K{\"o}hler et al. integrate segment-based shortest-path planning with MPC for mobile robots in cluttered environments \cite{kohler2025mpc}. JEPTERA uses JEPA-based UAV motion predictions to evaluate candidate actions within MPC, with PID providing low-level control.

\subsection{Predictive World Models for Planning}

Learning-based world models are used to estimate future states or representations, serving as a foundation for decision-making. Within the autonomous intelligence framework proposed by LeCun, JEPA learns to make predictions in a latent space, rather than reconstructing every detail of the input \cite{lecun2022path}. I-JEPA applies this principle to predict representations of specific target portions of an image based on available context \cite{assran2023ijepa}. Furthermore, V-JEPA extends this predictive learning approach to video data, yielding representations that encompass information regarding both appearance and motion \cite{bardes2024vjepa}.

The use of predictive representations for action-taking is further developed through V-JEPA 2, which incorporates an action-conditioned world model for robot planning \cite{assran2025vjepa2}. Meanwhile, DreamerV3 demonstrates the use of a world model to learn behaviors via reinforcement learning \cite{hafner2025dreamerv3}. Both approaches highlight the role of predictive models in decision-making, yet they employ different mechanisms for learning and action selection. JEPTERA positions JEPA as a model for predicting a UAV’s motion response to candidate actions, while action selection is performed via MPC.

\subsection{Safety-Aware Planning and Platform Integration}

A safety layer can complement a primary controller by checking or modifying actions based on safety constraints \cite{delecki2025diffusion}. Ames et al. formulated control based on Control Barrier Functions (CBF) using quadratic programming to integrate control objectives with safety requirements \cite{ames2017cbf}. Zeng et al. incorporated discrete-time CBFs into MPC, ensuring that safety constraints are considered throughout the prediction horizon \cite{zeng2021mpccbf}. Such approaches provide a foundation for incorporating safety as an explicit component of action selection.

Another approach is the predictive safety filter, which evaluates proposed actions using predicted dynamics and modifies them, if necessary, to satisfy constraints \cite{wabersich2021safetyfilter}. In JEPTERA, a safety check is implemented as an additional layer on top of the JEPA–MPC action selection process. The influence of this layer is evaluated through an ablation study to assess its impact on collisions, navigation success, and travel efficiency.

\section{Proposed Method}
\subsection{Architecture Overview}

At each planning cycle, the system updates voxel occupancy and obstacle tracks from the latest LiDAR observations. Route guidance and an observed free-space corridor support continuous optimization of the control sequence and path geometry. The optimizer uses locally linearized JEPA motion-effect predictions and a nominal response model. Optimized and backup plans are subsequently re-evaluated and checked for admissibility. An execution guard checks the held command between planning updates, while a platform adapter converts the resulting velocity reference into the controller interface.

\subsection{State and Environment Representation}

The translational state of the UAV used in planning consists of its position $\mathbf{p}_t \in \mathbb{R}^{3}$ and velocity $\mathbf{v}_t \in \mathbb{R}^{3}$, expressed in the world coordinate frame 
$
\mathbf{x}_t =
\begin{bmatrix}
\mathbf{p}_t & 
\mathbf{v}_t
\end{bmatrix}^\intercal
\in \mathbb{R}^{6}. 
\label{eq:planning_state}
$
Position and velocity are expressed in meters and meters per second, respectively. Information regarding the UAV's attitude is also used by the execution guard. Observations from each platform are converted into a common coordinate convention and set of units before being used by the planner.

The planning input $\mathbf{u}_{k|t} \in \mathbb{R}^{3}$ consists of a translational velocity reference:
\begin{equation}
\mathbf{u}_{k|t} =
\begin{bmatrix}
v^{\mathrm{ref}}_{x,k|t} &
v^{\mathrm{ref}}_{y,k|t} &
v^{\mathrm{ref}}_{z,k|t}
\end{bmatrix}^\intercal. 
\label{eq:velocity_reference}
\end{equation}
Here, $t$ denotes the current planning time, while $k = 0,\ldots,H-1$ represents the control step index within a horizon of $H$ steps. Each component specifies the desired velocity along a world coordinate axis, measured in meters per second. These references are continuous values and are distinct from motor commands. The selected commands undergo execution checks and rate-of-change limiting before being converted for the low-level controller interface.

Three dimensional occupancy mapping provides a representation of free, occupied, and unknown space to support robot navigation\cite{hornung2013octomap}. The environment is represented by a voxel occupancy map $\mathcal{M}_t$ and a set of obstacle tracks $\mathcal{T}_t$. The voxel map is constructed from LiDAR observations to distinguish between space that is observed to be occupied, observed to be free, or unknown. This representation is used to support route searching, corridor generation, and trajectory validation.

Separately, LiDAR observations over time are used to associate objects and estimate their motion. The set of tracks $\mathcal{T}_t$ contains estimates of the position, spatial extent, velocity, and uncertainty of obstacles. This information is used to predict the future positions of obstacles and to check for potential encounters with the UAV's trajectory. Tracking information is also used to flag dynamic areas within the voxel map.

\subsection{JEPA-Based Motion Prediction}

JEPA estimates the effect of a sequence of commands on UAV motion based on the observation context and action history. The prediction process is summarized as follows:
\begin{align}
\mathbf{z}_t &= E_\theta(\mathcal{O}_t), \label{eq:jepa_prediction} \\
\hat{\mathbf{z}}_{k|t}(\mathbf{U})
&= F_\phi^{(k)}(\mathbf{z}_t,\mathcal{A}_t,\mathbf{U}).
\end{align}

The encoder $E_\theta$ generates a latent representation from the observation context $\mathcal{O}_t$. The predictor $F_\phi^{(k)}$ produces a $k$-step-ahead prediction using this representation, the action history $\mathcal{A}_t$, and the sequence of velocity references $\mathbf{U}$.

During the predictor training phase, with the encoder kept frozen, the parameters $\phi$ are optimized by minimizing the mean absolute error in the latent space:
\begin{equation}
\mathcal{L}_{\mathrm{pred}}
=
\frac{1}{NHd}
\sum_{i=1}^{N}\sum_{k=1}^{H}
\left\| 
\hat{\mathbf{z}}_{i,k}
-\mathbf{z}^{\mathrm{target}}_{i,k}
\right\|_1,
\label{eq:jepa_prediction_loss}
\end{equation}
where $N$ is the number of samples, $H$ is the number of prediction steps, and $d$ is the number of latent components being compared. The target latent representations are obtained from target observations using the frozen encoder. The latent predictions are subsequently converted into estimates of displacement effects via a motion readout:
\begin{equation}
\widehat{\Delta\mathbf{p}}_{k|t}
=
\mathbf{W}^{\top}
\left(
\hat{\mathbf{z}}_{k|t}(\mathbf{U})
-
\hat{\mathbf{z}}_{k|t}(\mathbf{0})
\right). 
\label{eq:jepa_motion_readout}
\end{equation}
The matrix $\mathbf{W}$ maps the difference between the prediction and a zero-velocity reference into a displacement effect, expressed in meters. This estimate is used in conjunction with a nominal response model within an MPC framework. The movement of obstacles is estimated separately via LiDAR tracking.

\subsection{Model Predictive Planning}

MPC optimizes the sequence of speed references (U) and path-defining points (P) using SLSQP. The implementation employs four control steps with an interval of 0.2 seconds, resulting in a prediction horizon of 0.8 seconds. The cost function incorporates reference tracking, path length, motion smoothness, deviation from the route, and risk relative to obstacles:

\begin{equation}
\begin{aligned}
\min_{\mathbf{U},\mathbf{P}}\quad J
={}&C_{\mathrm{track}}(\mathbf{U})+8L(\mathbf{U,P})
+C_{\mathrm{smooth}}(\mathbf{U})\\
&+C_{\mathrm{route}}(\mathbf{U})+C_{\mathrm{risk}}(\mathbf{U,P})
+C_{\mathrm{static}}(\mathbf{U}).
\end{aligned}
\label{eq:mpc_objective}
\end{equation}
Here, $C_{track}$ measures the deviation of the predicted motion from the terminal reference. $L(P)$ represents the length of the projected path, $C_{smooth}$ penalizes sharp changes in motion, and $C_{route}$ measures the deviation from the guiding route. $C_{risk}$ imposes a penalty for proximity to corridor boundaries and predicted dynamic obstacles, while $C_{static}$ imposes a penalty for proximity to static occupied spaces.

\subsection{Safety Evaluation and Low-Level Control}

The safety check evaluates the UAV’s trajectory against voxel occupancy and predicted obstacle movement, including the response during braking. In the adaptive variant, the UAV’s continued motion prior to braking is also assessed, as coming to a stop could potentially lead to a dynamic conflict. During execution, a safety guard continuously monitors the active commands and can trigger braking if a risk is detected or if the UAV’s tilt exceeds a predefined limit.

Selected velocity references are subjected to rate-of-change constraints before being passed to the controller interface. In the PyBullet implementation, a cascaded PID controller calculates the required thrust and attitude based on position and velocity errors, subsequently determining motor RPMs through attitude control. Thus, the MPC scheme defines the motion references, while the PID controller handles their execution within the UAV’s dynamics.

\subsection{Platform-Agnostic Interface}

JEPTERA separates the planning core from simulator interfaces through platform adapters and YAML configurations. Input adapters normalize coordinate frames and units for UAV states and sensor observations, while output adapters translate velocity references into platform-specific control commands. YAML configurations define vehicle geometry, motion limits, response parameters, and velocity conversions. This separation enables integration across simulators without modifying the core planning logic.

\section{Experimental Setup}
\subsection{Simulation Environments}

\begin{figure}[t]
    \centering
    \includegraphics[
    width=1.0\linewidth,
    trim=0 0.2cm 0 0.2cm,
        clip
    ]
    {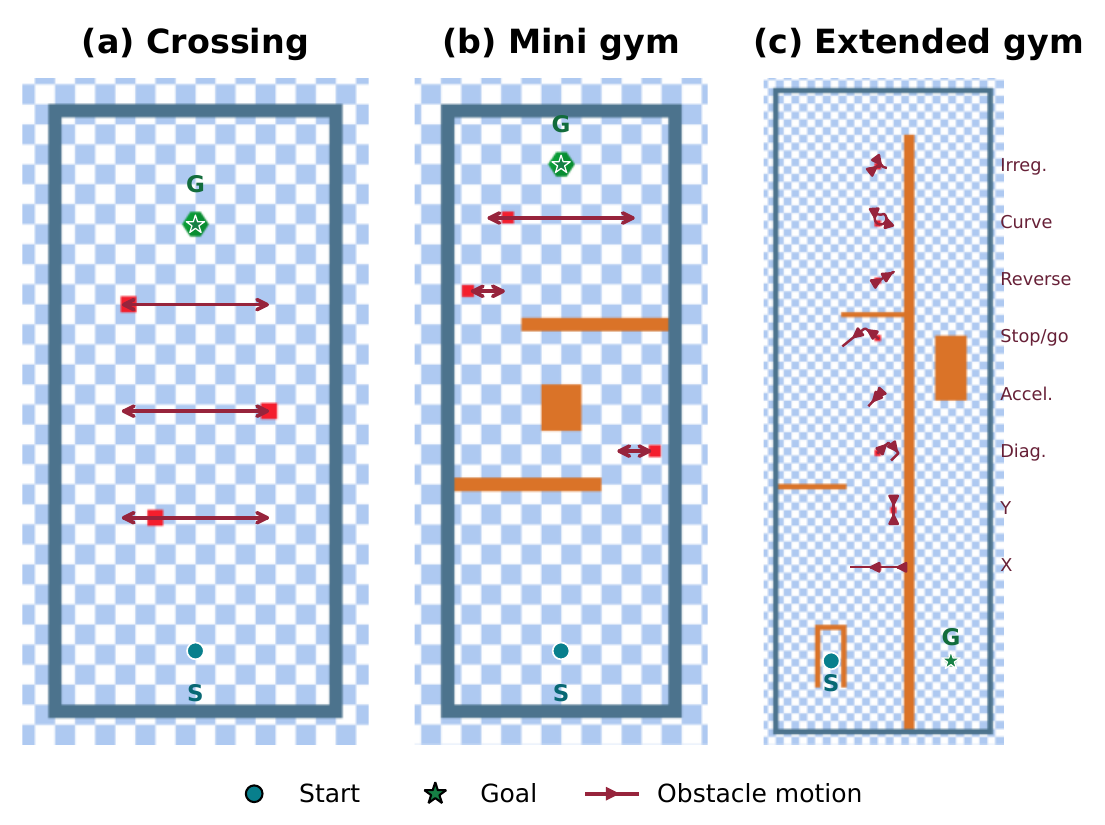}
    \caption{Layouts of the three gym scenarios used for the ablation study}
    \label{fig:gym_env}
\end{figure}

The evaluation encompasses testing navigation capabilities and cross-simulator portability. Navigation testing uses three gym scenarios featuring both static and dynamic obstacles within gym-pybullet-drones \cite{panerati2021learning}. The scenario layouts are shown in Fig.~\ref{fig:gym_env}

Navigation is additionally evaluated on the City A, City B, and City C maps to assess performance across different spatial layouts. Portability testing involves MuJoCo, Gazebo, and CARLA through adapters and platform-specific configurations while preserving the core planning logic \cite{mujocoDronesGym,koenig2004gazebo,dosovitskiy2017carla}.

\subsection{Compared Methods and Evaluation Protocol}

Six configurations were compared to evaluate the contributions of the architectural components: MPC–PID, MPC–PID with a safety layer, JEPA–PID, JEPA–PID with a safety layer, JEPA–MPC–PID, and JEPA–MPC–PID with a safety layer. Each configuration was tested across three gym scenarios using ten replications per scenario, resulting in a protocol comprising 180 runs. The same scenario-replication pairs were used for all configurations to ensure a fair comparison.

In the MPC–PID configuration, motion prediction uses a nominal response model without the contribution of JEPA. The JEPA–PID configuration selects command candidates based on the predicted motion effects from JEPA, without performing a joint optimization of the control sequence and path geometry. The JEPA–MPC–PID configuration integrates JEPA-based predictions with MPC optimization. Each configuration is tested both with and without an additional safety layer.

All outcomes, including collisions and failures to reach the goal, were incorporated into the evaluation. Testing across city maps used the complete JEPA–MPC–PID architecture with a safety layer to assess navigation capabilities across different spatial layouts. Cross-simulator testing was conducted separately to evaluate the portability of the implementation.

\subsection{Evaluation Metrics}

Navigation performance is evaluated based on success rate, collision rate, completion time, path length, and the planner's computation time. The success rate is calculated as the proportion of runs that reach the goal and complete the landing without any collisions, while the collision rate represents the proportion of runs that involve contact with obstacles. Completion time is measured using simulation time, whereas the planner's computation time is reported separately. Completion time and path length are compared based on successful runs, with the total number of successful runs also presented to ensure that failures are not overlooked.

\subsection{JEPA Training Setup}
The JEPA model is trained using sequences of voxel observations and UAV states, along with a history of control commands. The configuration employs eight context observations at 0.2-second intervals, a voxel resolution of 1 meter, and a latent dimension of 64. Following the representation pre-training phase, the encoder is frozen, and the predictor is trained for ten epochs using an L1 loss across 1,365 training windows and 735 validation windows. The model incorporating actual actions is compared against a zero-action model and a persistence baseline to evaluate the impact of action information on predictions.

Code, simulator adapters, and ablation instructions are available at \href{https://anonymous.4open.science/r/JEPTERA-33BE/}
{{\bfseries anonymous.4open.science/r/JEPTERA-33BE}}.

\section{Results and Discussion}

\subsection{JEPA Prediction Performance}

The model incorporating actual actions achieves an L1 latent error of 0.05343, compared to 0.05349 for the zero-action model and 0.06624 for the persistence baseline. These results indicate an error reduction of approximately 19.3\% relative to the persistence baseline, but only about 0.11\% relative to the zero-action model. Thus, the model’s predictions outperform the persistence baseline. However, the marginal difference between the model using actual actions and the one using zero actions suggests that the benefit of action information for the aggregate latent error remains limited. Its impact on navigation is evaluated through an ablation study in the following subsection.

\subsection{Navigation and Ablation Results}

Table~\ref{tab:navigation_efficiency} summarizes navigation success and performance. All failed runs ended in a collision.

The complete JEPA–MPC–PID architecture with a safety layer achieved the highest aggregate success, completing 28 of 30 trials. In the extended gym, it completed nine trials, compared with two without the additional safety layer. In the mini gym, however, MPC–PID with safety completed all trials, while the complete architecture experienced one collision. Adding safety also increased JEPA–PID success from one to 25 trials.

\begin{table*}[!t]
\centering
\caption{Navigation efficiency across the three gym scenarios}
\label{tab:navigation_efficiency}
\small
\setlength{\tabcolsep}{8pt}
\resizebox{\linewidth}{!}{%
\begin{tabular}{llcccc}
\hline
Scenario & Configuration & Success ${\uparrow}$ &
Time (s) ${\downarrow}$ & Path length (m) $\downarrow$ & Latency (ms) $\downarrow$ \\
\hline
Crossing
& MPC--PID & 10/10
& $99.23 \pm 3.54$ & $102.55 \pm 3.19$ & $192.94 \pm 2.85$ \\
& MPC--PID + Safety & 10/10
& $101.57 \pm 7.60$ & $102.87 \pm 7.37$ & $206.27 \pm 6.23$ \\
& JEPA--PID & 0/10
& --- & --- & $\mathbf{116.50 \pm 5.80}$ \\
& JEPA--PID + Safety & 10/10
& $\mathbf{92.52 \pm 3.44}$ & $\mathbf{99.13 \pm 2.75}$ & $231.32 \pm 11.06$ \\
& JEPA--MPC--PID & 10/10
& $97.98 \pm 4.19$ & $100.48 \pm 3.53$ & $232.73 \pm 3.31$ \\
& JEPA--MPC--PID + Safety & 10/10
& $101.03 \pm 7.58$ & $100.15 \pm 5.56$ & $245.63 \pm 5.29$ \\
\hline
Mini gym
& MPC--PID & 8/10
& $110.06 \pm 2.95$ & $124.17 \pm 6.41$ & $180.70 \pm 1.86$ \\
& MPC--PID + Safety & 10/10
& $114.90 \pm 9.54$ & $125.00 \pm 5.71$ & $190.53 \pm 4.56$ \\
& JEPA--PID & 0/10
& --- & --- & $\mathbf{89.13 \pm 3.12}$ \\
& JEPA--PID + Safety & 9/10
& $\mathbf{105.36 \pm 2.98}$ & $124.06 \pm 3.00$ & $146.55 \pm 20.23$ \\
& JEPA--MPC--PID & 10/10
& $111.69 \pm 5.35$ & $\mathbf{123.17 \pm 5.66}$ & $211.79 \pm 1.41$ \\
& JEPA--MPC--PID + Safety & 9/10
& $119.81 \pm 10.28$ & $126.01 \pm 2.13$ & $218.70 \pm 4.58$ \\
\hline
Extended gym
& MPC--PID & 4/10
& $282.35 \pm 33.69$ & $617.43 \pm 114.85$ & $183.91 \pm 4.85$ \\
& MPC--PID + Safety & 7/10
& $282.56 \pm 31.41$ & $569.19 \pm 80.83$ & $195.32 \pm 8.26$ \\
& JEPA--PID & 1/10
& $211.87$ & $556.61$ & $\mathbf{112.09 \pm 6.47}$ \\
& JEPA--PID + Safety & 6/10
& $\mathbf{211.05 \pm 7.29}$ & $\mathbf{548.38 \pm 8.17}$ & $201.69 \pm 11.85$ \\
& JEPA--MPC--PID & 2/10
& $267.30 \pm 11.17$ & $573.66 \pm 16.84$ & $225.37 \pm 4.73$ \\
& JEPA--MPC--PID + Safety & \textbf{9/10}
& $289.32 \pm 16.69$ & $549.47 \pm 7.27$ & $238.54 \pm 8.19$ \\
\hline
\end{tabular}%
}
\end{table*}

In addition to navigation success, efficiency is evaluated based on completion time, path length, and decision-making latency. Completion time and path length are summarized only from successful runs, whereas latency accounts for all runs. Full details of the results can be found in Table~\ref{tab:navigation_efficiency}.

\subsection{Navigation Across Urban Maps}

\begin{figure*}[!htbp]
    \centering
    \includegraphics[
    width=0.93\textwidth,
    trim=0 0.2cm 0 0.3cm,
        clip]
    {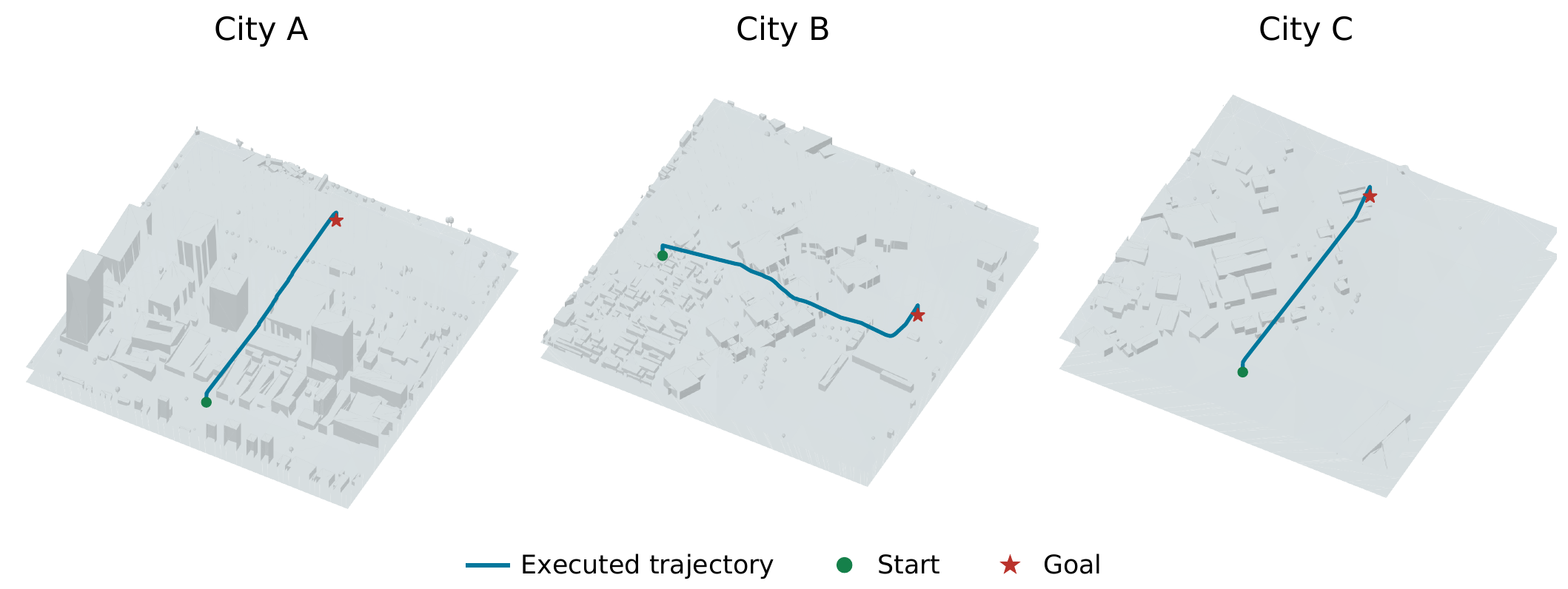}
    \caption{\centering Executed UAV trajectories across city environments}
    \label{fig:urban3d}
\end{figure*}

JEPTERA successfully completed the route on all three maps, reaching the landing point without any collisions. At City B, the trajectory shows changes in direction to navigate the airspace around buildings, whereas the paths taken in the City A and CityC environments are relatively direct. The metrics for each test are presented in Table~\ref{tab:urban_results}.

\begin{table}[H]
\centering
\caption{Navigation results across urban maps}
\label{tab:urban_results}
\resizebox{\linewidth}{!}{%
\begin{tabular}{lrrrr}
\hline
Map & Success & Time (s) & Path (m) & Error (cm) \\
\hline
City A & 1/1 & 186.93 & 383.55 & 0.90 \\
City B     & 1/1 & 172.73 & 356.85 & 2.12 \\
City C   & 1/1 & 166.73 & 328.09 & 1.33 \\
\hline
\end{tabular}%
}
\end{table}

\subsection{Cross-Simulator Portability}

The portability of JEPTERA is evaluated through its integration with MuJoCo, Gazebo, and CARLA. Each integration employs an adapter to align observations and control commands, alongside a YAML configuration to adjust platform-specific characteristics, all without altering the core logic of the planner. The results of the portability testing are presented in Table~\ref{tab:portability}. All three integrations successfully met the targets by using the same planner core, with adjustments made to the adapters and platform configurations.

\subsection{Discussion and Limitations}
Ablation results indicate that the integration of JEPA, MPC, PID, and a safety layer achieves the highest aggregate success rate, although it does not yield the best performance across every scenario. The improvement in success rates following the addition of a safety layer highlights the critical role of collision risk assessment in informing action selection. Nevertheless, the occurrence of collisions suggests a need for better coordination between prediction, planning, and execution when navigating dynamic obstacles. Testing across three urban maps along with cross-simulator integration, demonstrates the applicability of the core planner through the use of adapters and platform specific configurations. As this evaluation is currently limited to a single route per map and separate portability tests, it does not yet fully capture consistency across all operational conditions. Future work will involve real-time testing and flight operations using physical hardware to assess the system's reliability beyond the simulation environment.

\begin{table}[H]
\centering
\caption{Cross-simulator portability results}
\label{tab:portability}
\resizebox{\linewidth}{!}{%
\begin{tabular}{llrr}
\hline
Integration & Outcome & Time (s) & Error (m) \\
\hline
MuJoCo & Goal reached & 7.63  & 0.048 \\
Gazebo / ROS 2 & Goal reached & 12.23 & 0.047 \\
CARLA & Goal reached & 8.23 & 0.277 \\
\hline
\end{tabular}%
}
\end{table}

\section{Conclusion}

Reliable UAV navigation in dynamic urban environments requires accurate motion prediction and safety aware planning. We propose JEPTERA, a UAV navigation architecture that integrates JEPA-based motion effect prediction, MPC planning, PID control, and a safety layer. In an evaluation consisting of 180 runs, covering six configurations and three simulated scenarios. The complete architecture achieved the highest aggregate success rate, completing 28 out of 30 runs, though it did not outperform in every single scenario. Testing across city environments demonstrated collision-free route completion in each case. Furthermore, decoupling the core planner via adapters and YAML configurations facilitates integration across different simulators. These results highlight the potential of combining learning-based prediction with planning and safety checks for UAV navigation. Future development will focus on enhancing robustness against dynamic obstacles, as well as real-time validation and hardware-based flight testing.

\bibliographystyle{IEEEtran}
\bibliography{references}

\end{document}